\documentclass[review]{elsarticle}

\usepackage{lineno,hyperref}
\usepackage[utf8]{inputenc}
\usepackage{tabulary}
\modulolinenumbers[1]

\journal{Computers \& Geosciences}

\begin{document}

\begin{frontmatter}

\title{Segmentation of nearly isotropic overlapped tracks in
       photomicrographs using successive erosions as watershed
       markers\footnote{\textit{Authorship statement:} A. F. de S.
       designed the algorithms, wrote the code, analyzed the data and
       wrote the paper. W. M. S. counted the tracks in the images,
       analyzed the data and wrote the paper. S. G. obtained the images
       from the samples, analyzed the data, wrote the paper and
       supervised the research project. L. R. supervised the research
       project.}}

\author[unicamp,tubaf]{Alexandre Fioravante de Siqueira\corref{corrauthor}}
\cortext[corrauthor]{Corresponding author. Phone: +55(19)3521-5362.}
\ead{siqueiraaf@gmail.com}

\author[unicamp]{Wagner Massayuki Nakasuga}
\ead{wamassa@gmail.com}

\author[unicamp]{Sandro Guedes}
\ead{sguedes@ifi.unicamp.br}

\author[tubaf]{Lothar Ratschbacher}
\ead{lothar.ratschbacher@geo.tu-freiberg.de}

\address[unicamp]{Departamento de Raios Cósmicos e Cronologia, IFGW,
                  University of Campinas}
\address[tubaf]{Institut für Geologie, TU Bergakademie Freiberg}

\begin{abstract}
The major challenges of automatic track counting are distinguishing tracks
and material defects, identifying small tracks and defects of similar size,
and detecting overlapping tracks. Here we address the latter issue using
WUSEM, an algorithm which combines the watershed transform, morphological
erosions and labeling to separate regions in photomicrographs.
WUSEM shows reliable results when used in photomicrographs presenting
almost isotropic objects. We tested this method in two datasets of diallyl
phthalate (DAP) photomicrographs and compared the results when counting
manually and using the classic watershed. The mean automatic/manual
efficiency ratio when using WUSEM in the test datasets is $0.97 \pm 0.11$.
\end{abstract}


\end{frontmatter}


\section{Introduction}

Solid state nuclear track detectors (SSNTD) are materials such as inorganic
crystals, plastics and glasses, known to record the path of charged
particles. There are several applications for SSNTD in nuclear science;
for instance, measurements of radon gas \cite{BROWN2011}, boron neutron
capture therapy \cite{SMILGYS2013}, and age determination by fission
track dating \cite{WAGNER1992, GOTO2011, IMAYAMA2012, TAKAGI2013,
YUGUCHI2017, NINOMIYA2017}.

When a charged particle collide in a SSNTD, the ionization and/or the
collision with atoms modify the path by which the particle go through.
This damage in the SSNTD structure is called latent track. This track
becomes visible under an optical microscope after a convenient etching
process \cite{FLEISCHER1964}, and these tracks can be counted.
Photomicrographs can also be used, which could improve the counting
accuracy \cite{MORIWAKI1998}.

Procedures for measuring and counting tracks are time-consuming and
involve practical problems, e.g. variation in observer efficiency
\cite{GLEADOW2009}. An automatic method based on image processing
techniques could increase the track counting rate and improve counting
reproducibility. However, separating elements in nontrivial images is
one of the hardest tasks in image processing \cite{GONZALEZ2007}.

Automatic systems for separating, counting or measuring tracks have been
studied for a while, and several solutions were presented (e.g.
\cite{AZIMIGARAKANI1977, BEER1982, PRICE1985, TRAKOWSKI1984, MASUMOTO1990,
WADATSUMI1990, FEWS1992, PETFORD1992, PETFORD1993, ESPINOSA1996, BOUKHAIR2000,
DOLLEISER2002, BEDOGNI2003, YASUDA2005, TAWARA2008, GLEADOW2009, DESIQUEIRA2014,
RUAN2014}. Still, the precision of automatic methods is not satisfactory
yet; an automatic analysis could need to be manually adjusted by the
operator, being more time consuming than the usual measure \cite{YASUDA2005,
ENKELMANN2012}. The major challenges to automatic track counting are
detecting overlapping tracks, distinguishing tracks and material defects
(e.g. surface scratches due to polishing), and identifying small tracks
and defects of comparable size in the background of photomicrographs
\cite{GLEADOW2009}.

To address the problem of identifying overlapping ion tracks in photomicrographs,
we propose an algorithm based on the watershed transform \cite{BEUCHER1979}
using morphological erosions \cite{SERRA1982} as markers. A similar method
was used to separate packings of ellipsoidal particles represented using
X-Ray tomography \cite{SCHALLER2013}. We tested this method in two datasets
of diallyl phthalate (DAP) photomicrographs, and used the results to relate
the incident track energy with the mean gray levels and mean diameter
products for each sample from the first dataset.

\section{Material and methods}

We employed the ISODATA threshold to obtain the binary images used in
our tests. The WUSEM algorithm is based on the following techniques: 1.
morphological erosion; 2. watershed transform, and 3. labeling. We
describe these algorithms in this Section.

\subsection{The ISODATA threshold}

The Iterative Self-Organizing Data Analysis Technique (A) (ISODATA)
threshold \cite{BALL1967, RIDLER1978} is an histogram-based method. It
returns a threshold that separates the image into two pixel classes,
where the threshold intensity is halfway between their mean intensities.

When applied for two classes, ISODATA is always convergent \cite{VELASCO1980};
in our case, these classes are the tracks (our regions of interest, ROI)
and the background. We used the algorithm \texttt{filters.threshold\_isodata},
implemented in scikit-image \cite{VANDERWALT2014}.

\subsection{Structuring elements and morphological erosion}

In morphological image processing, a structuring element is a matrix
representing a mask, or a shape, that is used to retrieve information
about the shapes in an input image \cite{SERRA1982}. Erosion, a basic
operation in image processing, uses a chosen structuring element to shrink
the border of all ROI in a binary image; the shrinkage factor is
correspondent to the structuring element size \cite{GONZALEZ2007}.

In our tests (Section \ref{sec:experimental}) we used disks as structuring
elements, since tracks in DAP are mostly round in shape. The algorithms
used were \texttt{morphology.disk} and \texttt{morphology.erosion},
contained in scikit-image.

\subsection{Watershed transform}

The watershed algorithm is a non-parametric method which defines the
contours as the watershed of the gradient modulus of the gray levels of
the input image, considered as a relief surface detection method
\cite{BEUCHER1979}.

In this algorithm, the input image is seen in a three-dimensional
perspective. Two dimensions correspond to spatial coordinates, and the
third represents the gray levels. In this interpretation, we consider
three kinds of points \cite{GONZALEZ2007}:

\begin{enumerate}
    \item Points in regional minima.
    \item Points where a drop of water would flow to a common minimum.
    \item Points where a drop of water would flow to different minima.
    The set of these points is named \textit{watershed line}.
\end{enumerate}

The aim of watershed algorithms is to find the watershed lines. The
method used in this paper is implemented in the function
\texttt{morphology.watershed}, from scikit-image.

\subsection{Labeling algorithm}

In image processing, the labeling algorithm labels connected regions of
a binary input image, according to the 2-connectivity sense: all eight
pixels surrounding the reference pixel. Pixels receive the same label
when they are connected and have the same value. We used the algorithm
\texttt{measure.label} from scikit-image, implemented as described in
\cite{FIORIO1996}.

\subsection{Watershed Using Successive Erosions as Markers (WUSEM)
algorithm}

Here we present the WUSEM (Watershed Using Successive Erosions as Markers)
algorithm, which combines morphological erosions, the watershed transform
and labeling algorithms to separate regions of interest (ROI) in binary
images. The WUSEM algorithm and its steps follow (Figure
\ref{fig:wusem_scheme}):

\begin{figure*}[htb]  
    \centering
    \includegraphics[width=1\textwidth]{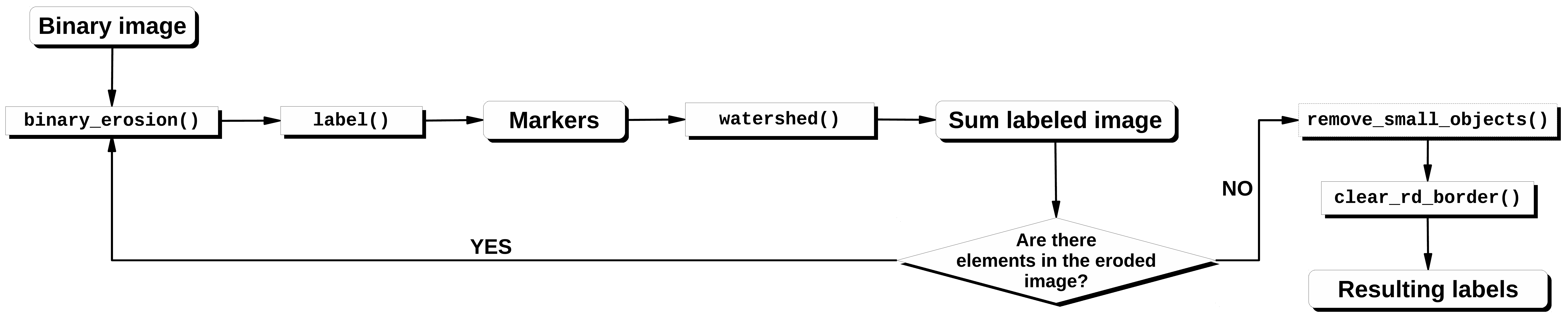}
    \caption{Representing WUSEM and the functions employed in its
    implementation.}
    \label{fig:wusem_scheme}
\end{figure*}

\begin{figure}[htb]  
    \centering
    \includegraphics[width=0.8\textwidth]{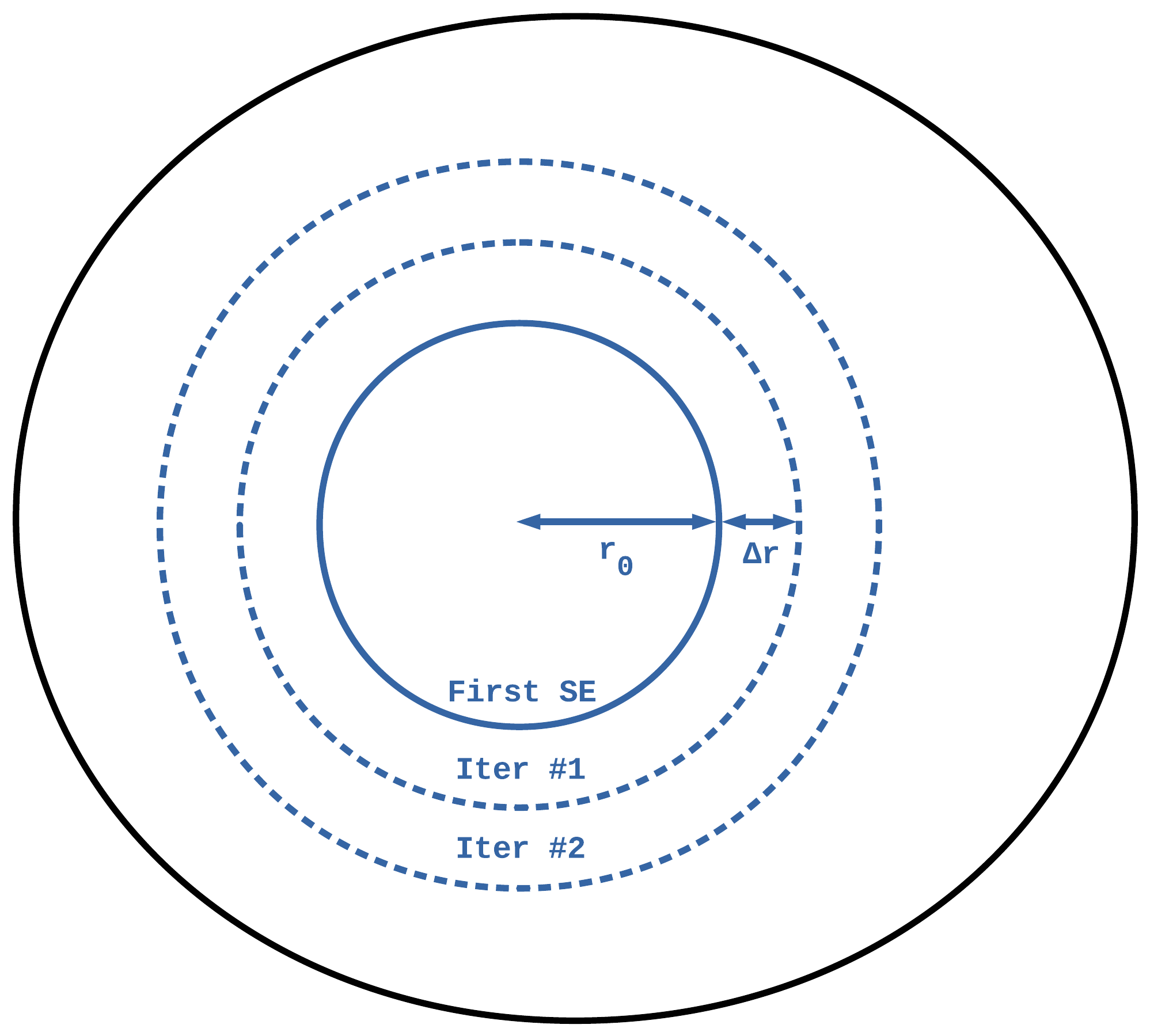}
    \caption{Representation of disks as structuring elements defined for
    each segmentation when employing WUSEM in an interest region. The
    radius of the first, second and third structuring elements are $r_0$,
    $r_0 + \Delta r$ and $r_0 + 2\Delta r$, respectively.}
    \label{fig:strel_representation}
\end{figure}

\begin{enumerate}
    \item The user define an initial radius ($r_0$) and an iterative radius
    ($\Delta r$) to a structuring element (Figure \ref{fig:strel_representation}).
    \item The input binary image is eroded using the structuring element
    with radius equal to $r_0$. This erosion is used as a marker to the
    watershed algorithm. Then, the resulting image is labeled. This is
    the first segmentation.
    \item A new structuring element is defined. Its radius is 
    $r_0 + \Delta r$. The input binary image is eroded with this new
    structuring element, the erosion is used as a marker to watershed,
    and the result is labeled. This is the second segmentation.
    \item The process continues until the eroded image does not have
    objects. Then, all segmentations are summed.
    \item The result is labeled again, to reorder the ROI, and labels
    with area smaller than 64 pixels are excluded, to ensure that noise
    will not affect the results. Tracks are counted according to the ``lower
    right corner'' method, where objects that touch the bottom and right
    edges are not counted. This leads to an accurate, unbiased measure
    \cite{RUSS2011}.
\end{enumerate}

\begin{figure}[htb]  
    \centering
    \includegraphics[width=0.8\textwidth]{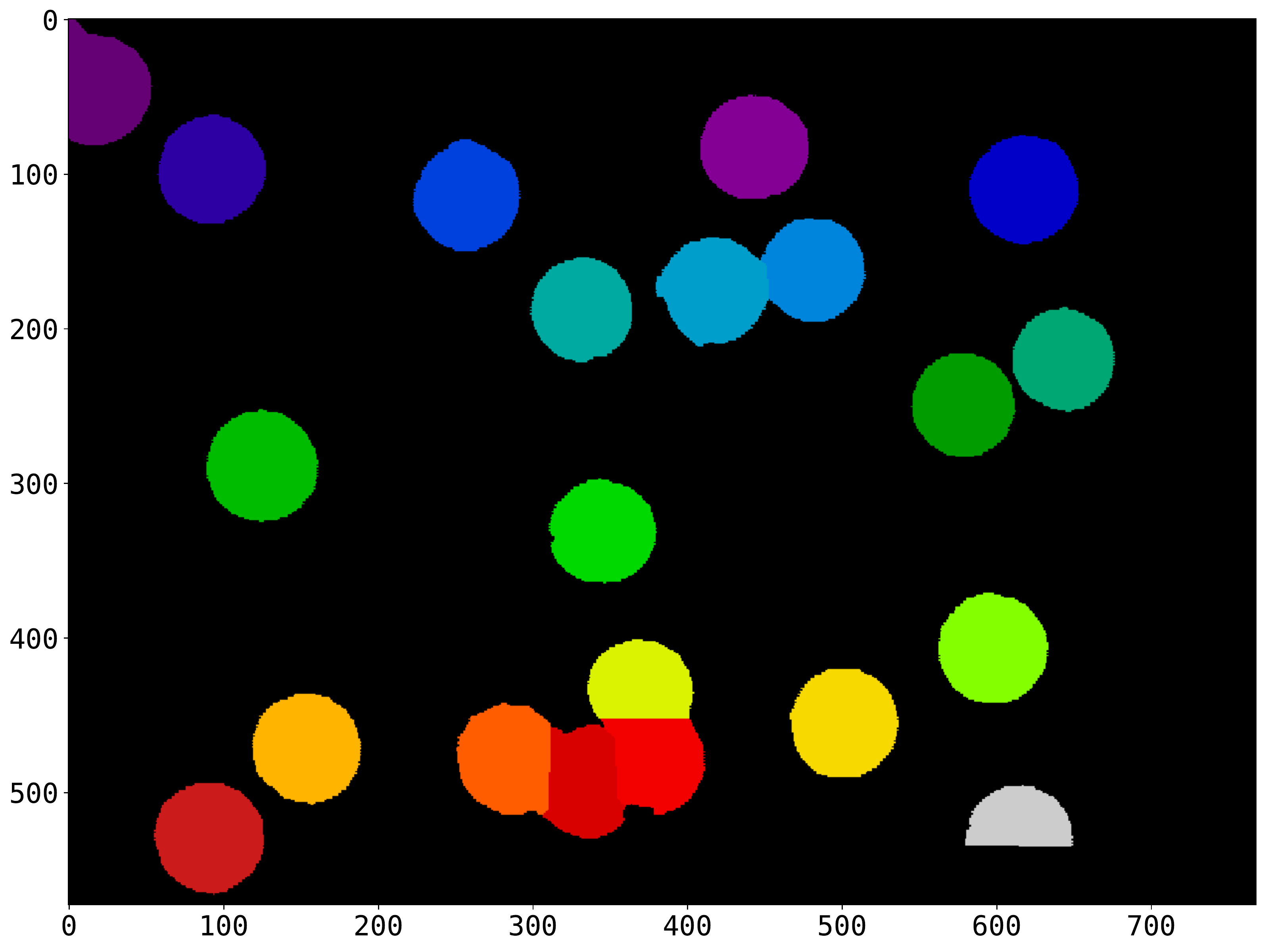}
    \caption{WUSEM algorithm application in an input image. First, the
    image is binarized using the ISODATA threshold. Then, the image is
    eroded using initial radius ($r_0$) and iterative radius ($\Delta r$)
    equal to 1 and 1 respectively, to ease visualization. The process
    continues until the eroded image has regions in it. All erosions are
    summed, and the result is labeled; then, regions with area smaller
    than 64 px are excluded. Each labeled track receives a color from
    the \texttt{nipy\_spectral} colormap. Finally, the function
    \texttt{enumerate\_objects()} is used to number the found tracks.
    Final results are shown to $r_0$ and $\Delta r$ equal to 10 and 4,
    respectively. Animation also available at
    \url{https://youtu.be/CtIOxhNISW8}.}
    \label{fig:wusem_animation}
\end{figure}

WUSEM is implemented in the function \texttt{segmentation\_wusem()}, available
in the Supplementary Material. It receives the arguments \texttt{str\_el},
\texttt{initial\_radius} and \texttt{delta\_radius}, representing the
structuring elements, $r_0$ and $\Delta r$, respectively. To exemplify
WUSEM’s capabilities, we used it to separate overlapping tracks within
the test photomicrographs (Figure \ref{fig:wusem_animation}).

\subsection{DAP photomicrographs}

We used two sets of diallyl phtalate (DAP, C$_{14}$H$_{14}$O$_{4}$)
photomicrographs to test the WUSEM algorithm. One of them was obtained
from detectors irradiated with $^{78}Kr$ tracks, and the other has
induced fission tracks.

\subsubsection{$^{78}Kr$ tracks}

The first dataset contains 362 photomicrographs of tracks from nine
different DAP plaques irradiated with $^{78}Kr$ ions at a nominal fluence of
$1.4\times 105\,cm^{-2}$, from a beam perpendicular to the detector surfaces
at GSI, Darmstadt, Germany\footnote{These photomicrographs are contained
in the folder \texttt{orig\_figures/dataset\_01}, available in the
Supplementary Material.}.

During the irradiation, detectors were covered with aluminum foils forming
a moderation layers of thicknesses varying from zero (no cover) to $90\,\mu m$.
Detectors were named after their cover thicknesses (K0, K20, K30, ..., K90).
Ions arrived at the setup with initial energy of 865 MeV and were slowed
down in the aluminum cover before hitting the detector surface. Incidence
energies were calculated using the software SRIM \cite{ZIEGLER2010}, and varied
from  18 MeV ($90\,\mu m$ cover) to  865 MeV (no cover). Detectors were etched
in a PEW solution (7.5 g KOH, 32.5 g ethanol, 10 g water) solution for
$4.5\pm0.2$ min at $65\pm3\,^{\circ} C$.

\begin{figure}[htb]  
    \centering
    \includegraphics[width=0.8\textwidth]{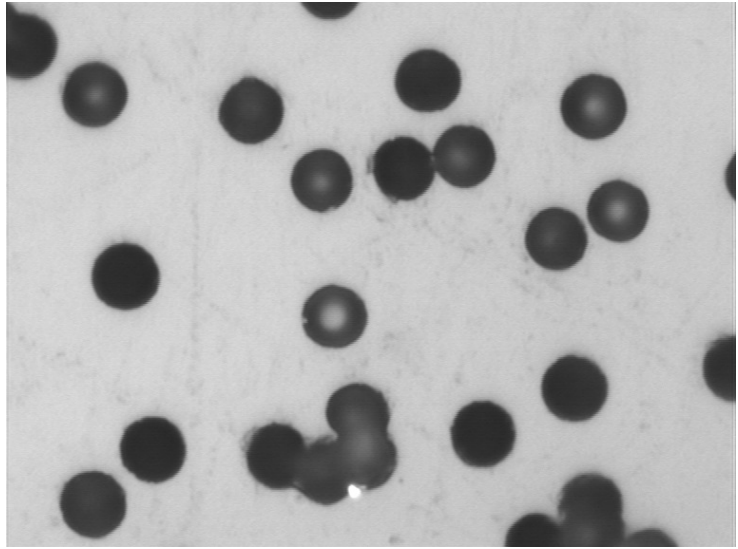}
    \caption{Photomicrograph from the first dataset, presenting tracks
    in a DAP sample.}
    \label{fig:test_figure1}
\end{figure}

Images were captured with a CCD camera coupled with a Zeiss microscope,
in reflected light mode, under 1250 $\times$ nominal magnification. Then,
the detectors were further etched for 4 minutes, total of $8.5\pm0.3$ min,
and new images were captured. This way, we obtained 18 subsets of images.
Tracks within these photomicrographs are almost isotropic and have the
same orientation, resembling circles (Figure \ref{fig:test_figure1}).

\subsubsection{Induced fission tracks}

\begin{figure*}[htb]  
    \centering
    \includegraphics[width=0.8\textwidth]{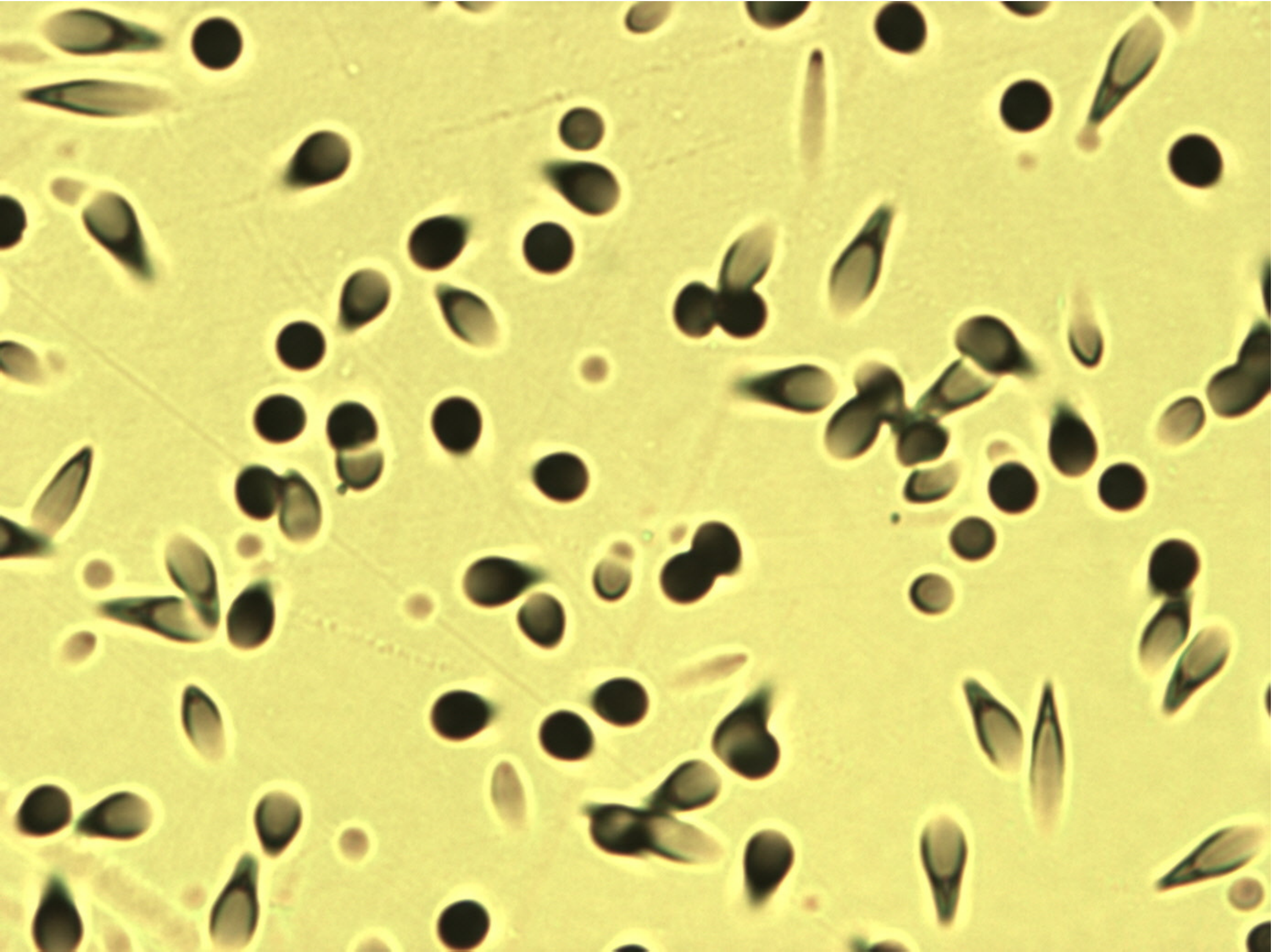}
    \caption{Input test photomicrograph from the second dataset,
    presenting tracks in a DAP sample.}
    \label{fig:test_figure2}
\end{figure*}

The second dataset contains 19 photomicrographs with two different magnifications
from a DAP plaque used as external detector, coupled to an apatite sample
and irradiated with thermal neutrons to induce fission in the $^{235}U$
atoms inside the mineral\footnote{These photomicrographs are contained
in the folder \texttt{orig\_figures/dataset\_02}, available in the
Supplementary Material.}. During the fission process, two fragments are
released and, eventually, are detected by the DAP plaque. Fragments
arrive at the detector surface with different energies and incidence
angles, resulting in a variety of track formats. Hence, counting tracks
in these photomicrographs is more complex than in the previous case
(Figure \ref{fig:test_figure2}).

\subsection{License and reusability}

The WUSEM algorithm and several functions for its implementation lie
within the packages Numpy \cite{VANDERWALT2011}, Scipy \cite{JONES2001},
Matplotlib \cite{HUNTER2007}, scikit-image \cite{VANDERWALT2014}, among
others. All code published with this paper is written in Python 3
\cite{VANROSSUM1995} and is available under the GNU GPL v3.0 license, and
all photomicrographs and figures distributed with this paper are available
under the CC-BY 2.0 license.

\section{Experimental}
\label{sec:experimental}

\subsection{Processing photomicrographs of $^{78}$Kr tracks}

\subsubsection{Exemplifying the methodology}

\begin{figure}[htb]  
    \centering
    \includegraphics[width=1\textwidth]{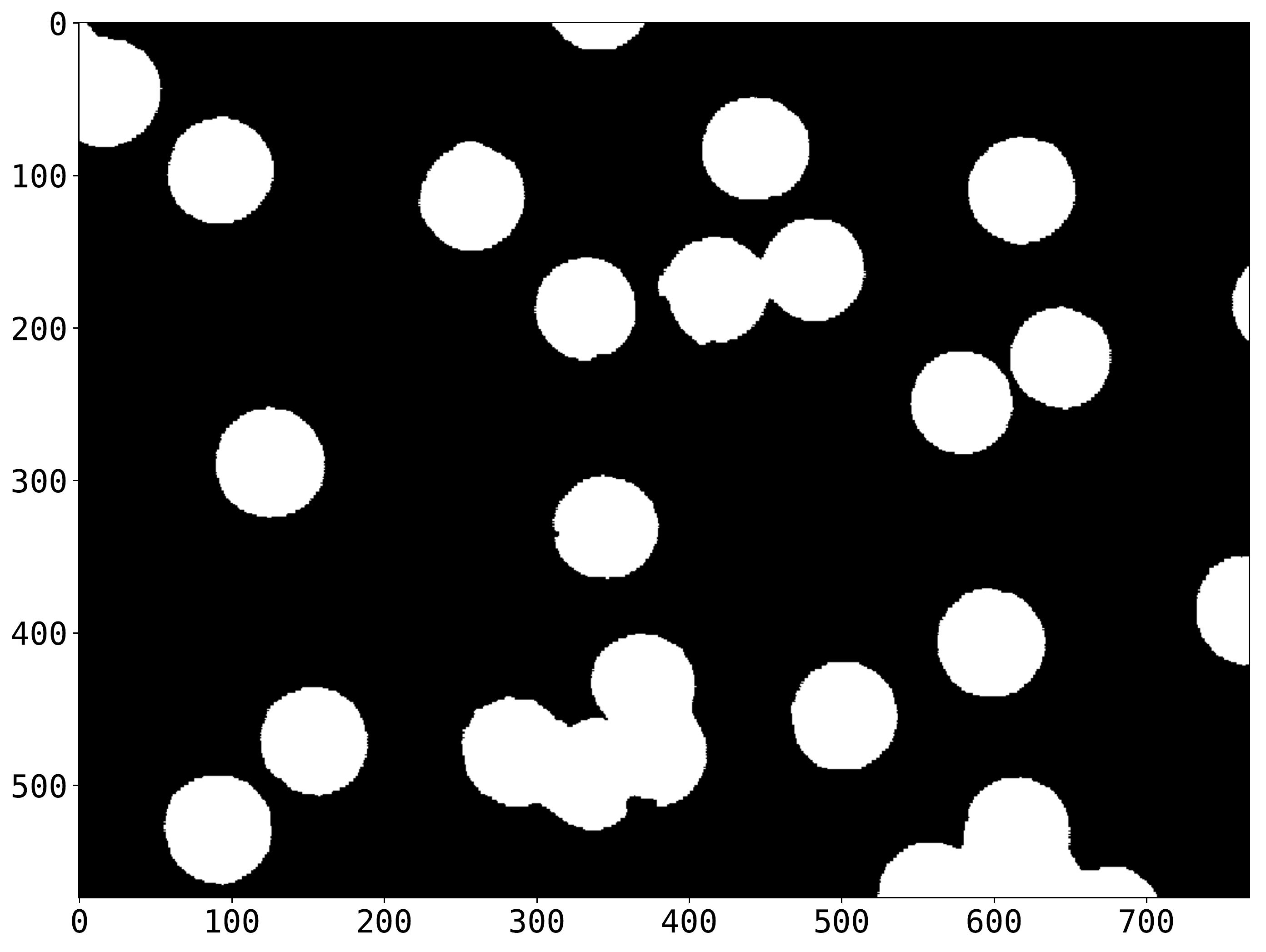}
    \caption{Input photomicrograph (Figure \ref{fig:test_figure1})
    binarized using the ISODATA threshold (threshold = 128) and region
    filling. Colormap: \texttt{gray}.}
    \label{fig:test_binary1}
\end{figure}

Here we use a photomicrograph from the first dataset\footnote{Image
\texttt{K90\_incid4,5min\_3.bmp} from the folder
\texttt{orig\_figures/dataset\_01/Kr-78\_4,5min/K90\_incid}, available
in the Supplementary Material.} to exemplify WUSEM (Figure
\ref{fig:test_figure1}). We binarized this photomicrograph using the
ISODATA threshold \cite{BALL1967, RIDLER1978} (Figure
\ref{fig:test_binary1}). Different gray levels in some tracks may not be
properly separated, complicating the extraction of track features. To
address this issue, we filled the regions in the binary image using the
function \texttt{ndimage.morphology.binary\_fill\_holes()} from scipy. 

\begin{figure}[htb]  
    \centering
    \includegraphics[width=0.7\textwidth]{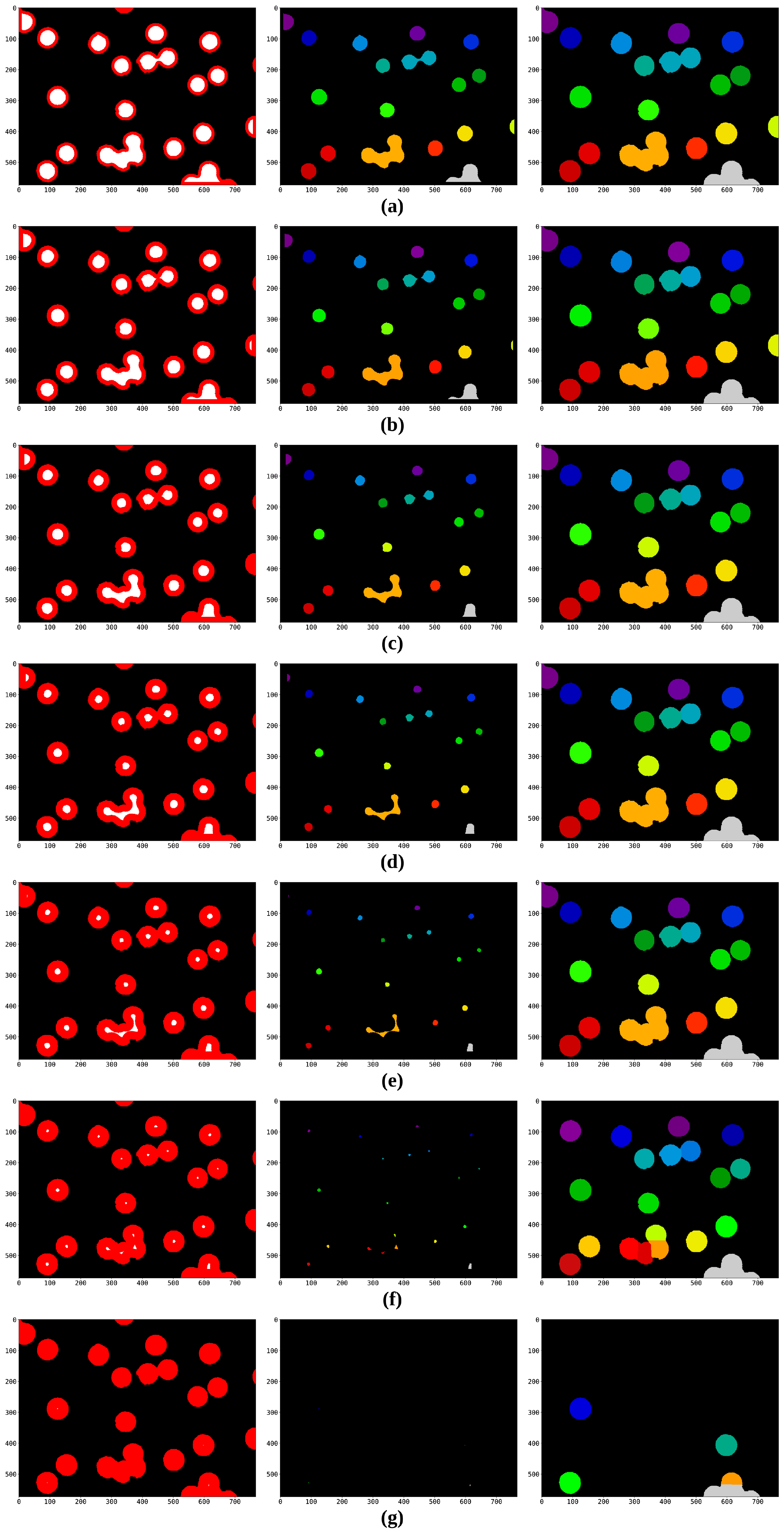}
    \caption{Processing an input image using WUSEM. Each line presents a
    segmentation step. Left column: original binary image (red) and erosion
    obtained according to the variation of \textit{delta\_radius} (white).
    Center column: erosion results labeled, used as markers. Right column:
    watershed results when using generated markers. Parameters for WUSEM
    algorithm: \texttt{initial\_radius = 10}, \texttt{delta\_radius = 4}.}
    \label{fig:wusem_processing}
\end{figure}

After binarizing the input image, the WUSEM algorithm separates the
overlapping tracks (Figure \ref{fig:wusem_processing}). For this example,
we chose \texttt{initial\_radius = 10} and \texttt{delta\_radius = 4} as
parameters.

\begin{figure}[htb]  
    \centering
    \includegraphics[width=1\textwidth]{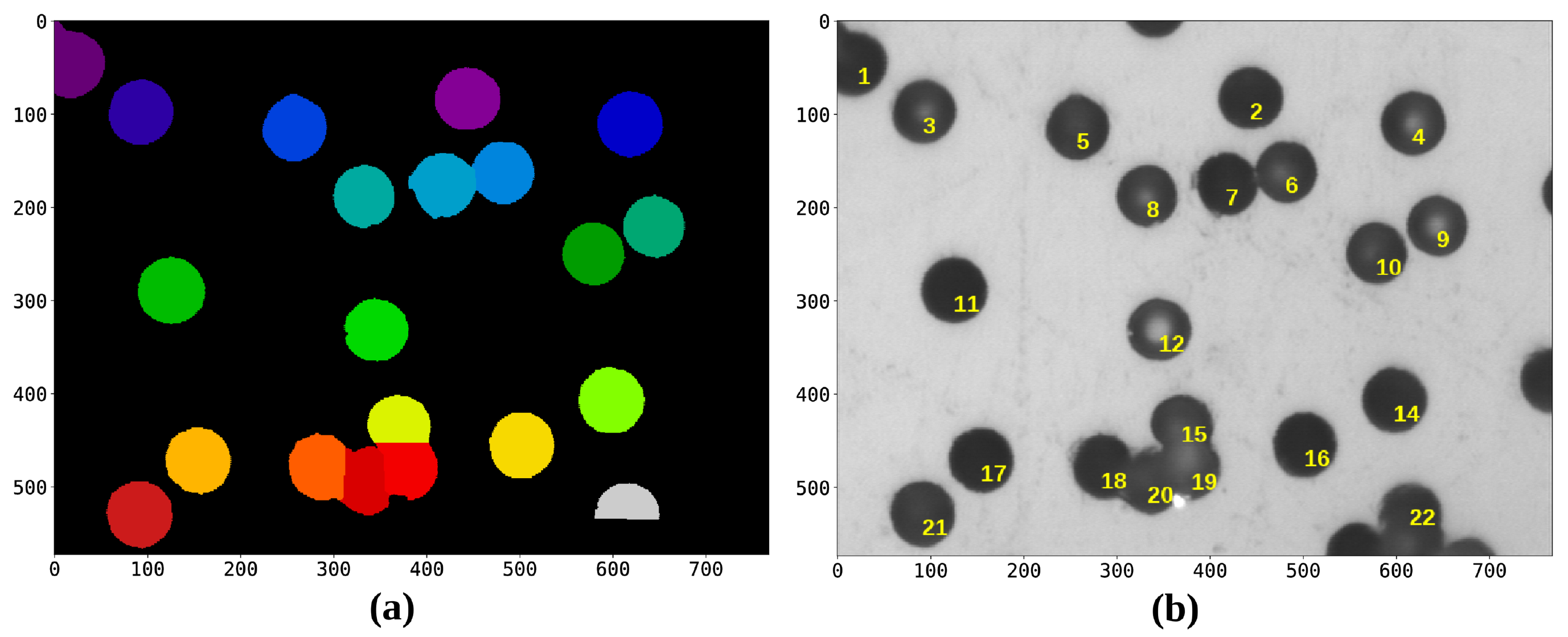}
    \caption{\textbf{(a)} Labels generated from tracks in Figure
    \ref{fig:test_figure1} using the WUSEM algorithm. \textbf{(b)} Tracks
    in (a) enumerated using \texttt{enumerate\_objects()}. Tracks in the
    lower or right corners are not counted, according to the ``lower right
    corner'' method. Parameters for WUSEM algorithm:
    \texttt{initial\_radius = 10}, \texttt{delta\_radius = 4}. Colormaps:
    \textbf{(a)} \texttt{nipy\_spectral}, \textbf{(b)} \texttt{gray}.}
    \label{fig:enumerated_result}
\end{figure}

Tracks were counted using the ``lower right corner'' method, i.e., border
tracks are only counted if they lie on the right or bottom edges of the
image. We used the function \texttt{clear\_rd\_border()}, given in the
Supplementary Material, to remove the tracks in these edges. WUSEM returns
a labeled region, which can be used as parameter to the function
\texttt{enumerate\_objects()} (Figure \ref{fig:enumerated_result}).

\subsubsection{Comparison between manual and automatic counting}

An experienced observer can easily distinguish tracks in the photomicrograph,
even when several tracks are superimposed. For this reason, manual counting
is considered the control in the comparison with the automatic counting.
In the following, the WUSEM algorithm is applied to the photomicrograph
set and the processing parameters are studied.

\begin{figure}[htb]  
    \centering
    \includegraphics[width=1\textwidth]{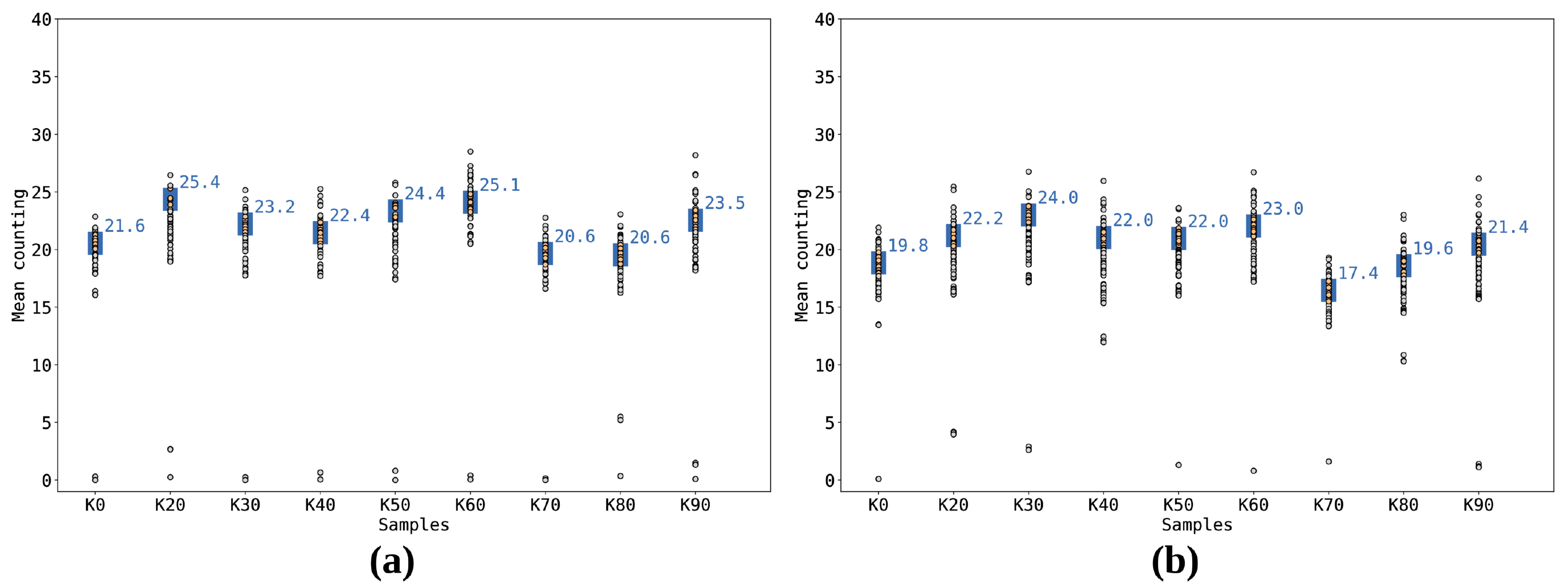}
    \caption{Manual counting mean (top of the blue bar; values on the
    right) for each sample and automatic counting results with mean
    within (orange points) and outside (gray points) the tolerance
    interval (blue bar) for the first dataset.}
    \label{fig:tolerance_int}
\end{figure}

We established an arbitrary value of up to two tracks less than the mean
of the manual counting as a tolerance. WUSEM’s parameters become candidates
if the automatic counting lies within the tolerance interval, i.e.
$0 < \mu_{n_{manual}} – n_{auto} < 2$, where $n$ is the track number
obtained by each approach (Figure \ref{fig:tolerance_int}).

WUSEM’s best parameters for this study are the ones within the tolerance
interval for most samples. According to the stated comparison, the best
parameters are \texttt{initial\_radius = 10}, \texttt{delta\_radius = 20}
for 4.5 min samples, and \texttt{initial\_radius = 10},
\texttt{delta\_radius = 14} for 8.5 min samples.

\begin{figure*}[htb]  
    \centering
    \includegraphics[width=0.8\textwidth]{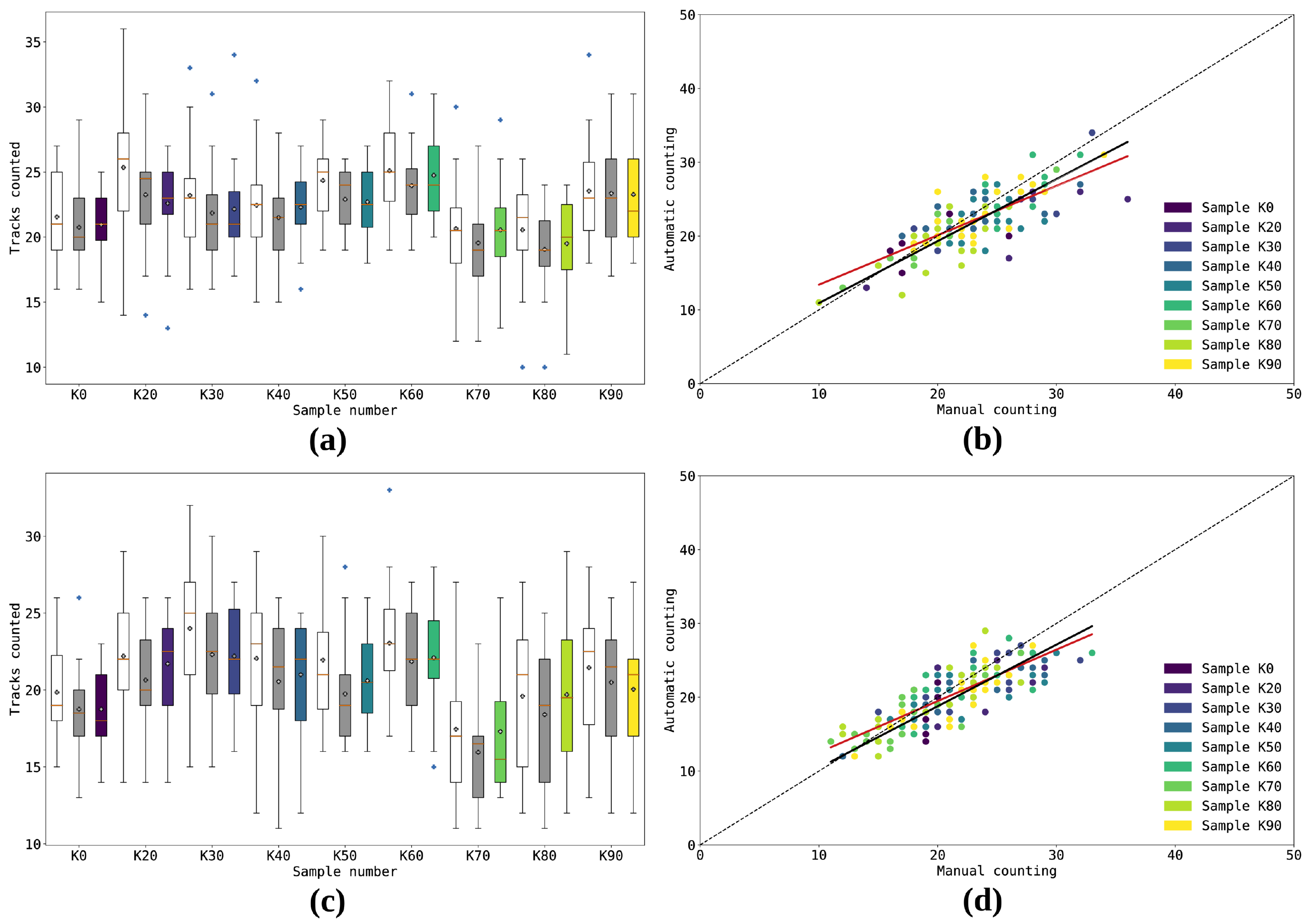}
    \caption{Comparison between manual and automatic counting for \textbf{(a, b)}
    4.5 min etching samples and \textbf{(c, d)} 8.5 min etching samples.
    \textbf{(a, c)} white: manual counting. Gray: flooding watershed counting. Red
    line: distribution median. White signal: distribution mean. \textbf{(b, d)}
    dashed: 1:1 line. Red line: regression for the WUSEM counting data.
    Black line: regression for the flooding watershed counting data.}
    \label{fig:stats_set1}
\end{figure*}

Using the best parameters defined, we compared manual, classic (flooding)
watershed and WUSEM counting for each sample (Figure \ref{fig:stats_set1},
Table \ref{tab:stats_set1}). The classic watershed algorithm used is
implemented in ImageJ \cite{RUEDEN2017}, and the full method consisted in
binarizing the input image, applying the watershed and removing small
objects. The source code for these operations and instructions on how to
use it are given in the Supplementary Material.

Since tracks in this dataset have the same shapes, we can attribute an
efficiency of 100 \% for manual counting. WUSEM counting was initially
set to obtain a smaller number of tracks when compared to the manual
counting. However, WUSEM counting returns false positives, i.e.,
incorrectly labels background regions as ROI (points above the 1:1 line
in Figure \ref{fig:stats_set1} (b) and (d)). To avoid false positives,
one could use a more restrictive criterion, such as eccentricity (Section
\ref{sec:ion_energies}).

\begin{table*}[h]  
\centering
    {\scriptsize
    \begin{tabulary}{1\textwidth}{CCCCCCC}
    \hline
    & \multicolumn{2}{c}{\textbf{Manual counting ($\mu\pm\,1\sigma$)}} & \multicolumn{2}{c}{\textbf{WUSEM counting ($\mu\pm\,1\sigma$)}} & \multicolumn{2}{c}{\textbf{Efficiency $\pm\,1\sigma$}} \\
    \hline
    \textbf{Sample} & \textbf{4.5 min} & \textbf{8.5 min} & \textbf{4.5 min} & \textbf{8.5 min} & \textbf{4.5 min} & \textbf{8.5 min} \\
    \hline
    \textbf{K0} & $22 \pm 3$ & $20 \pm 3$ & $21 \pm 2$ & $19 \pm 3$ & $0.99 \pm 0.12$ & $0.95 \pm 0.11$ \\
    \textbf{K20} & $25 \pm 5$ & $22 \pm 4$ & $23 \pm 3$ & $22 \pm 3$ & $0.90 \pm 0.11$ & $0.99 \pm 0.14$ \\
    \textbf{K30} & $23 \pm 4$ & $24 \pm 4$ & $22 \pm 4$ & $22 \pm 3$ & $0.96 \pm 0.12$ & $0.94 \pm 0.10$ \\
    \textbf{K40} & $22 \pm 4$ & $22 \pm 4$ & $22 \pm 3$ & $21 \pm 3$ & $1.01 \pm 0.10$ & $0.96 \pm 0.08$ \\
    \textbf{K50} & $24 \pm 2$ & $22 \pm 4$ & $23 \pm 3$ & $20 \pm 3$ & $0.94 \pm 0.11$ & $0.95 \pm 0.13$ \\
    \textbf{K60} & $25 \pm 3$ & $23 \pm 4$ & $25 \pm 3$ & $22 \pm 3$ & $0.99 \pm 0.08$ & $0.97 \pm 0.13$ \\
    \textbf{K70} & $21 \pm 4$ & $17 \pm 4$ & $21 \pm 4$ & $17 \pm 4$ & $1.00 \pm 0.07$ & $1.00 \pm 0.13$ \\
    \textbf{K80} & $21 \pm 4$ & $20 \pm 5$ & $20 \pm 4$ & $20 \pm 4$ & $0.96 \pm 0.13$ & $1.02 \pm 0.14$ \\
    \textbf{K90} & $24 \pm 4$ & $21 \pm 4$ & $23 \pm 4$ & $21 \pm 4$ & $0.99 \pm 0.11$ & $0.94 \pm 0.09$ \\
    \hline
    \end{tabulary}
    }
    \caption{Mean, standard deviation, and automatic/manual efficiency
    ratio for each sample of the first dataset.}
    \label{tab:stats_set1}
\end{table*}

Counting reproducibility is important for considering the reliability of
track counting. Despite the variations in track characteristics, the
efficiencies of WUSEM’s automatic counting remained constant within
uncertainties, when compared to manual counting.

\subsubsection{Relating ion energies with diameter product and mean gray
levels}
\label{sec:ion_energies}

\begin{figure}[htb]  
    \centering
    \includegraphics[width=1\textwidth]{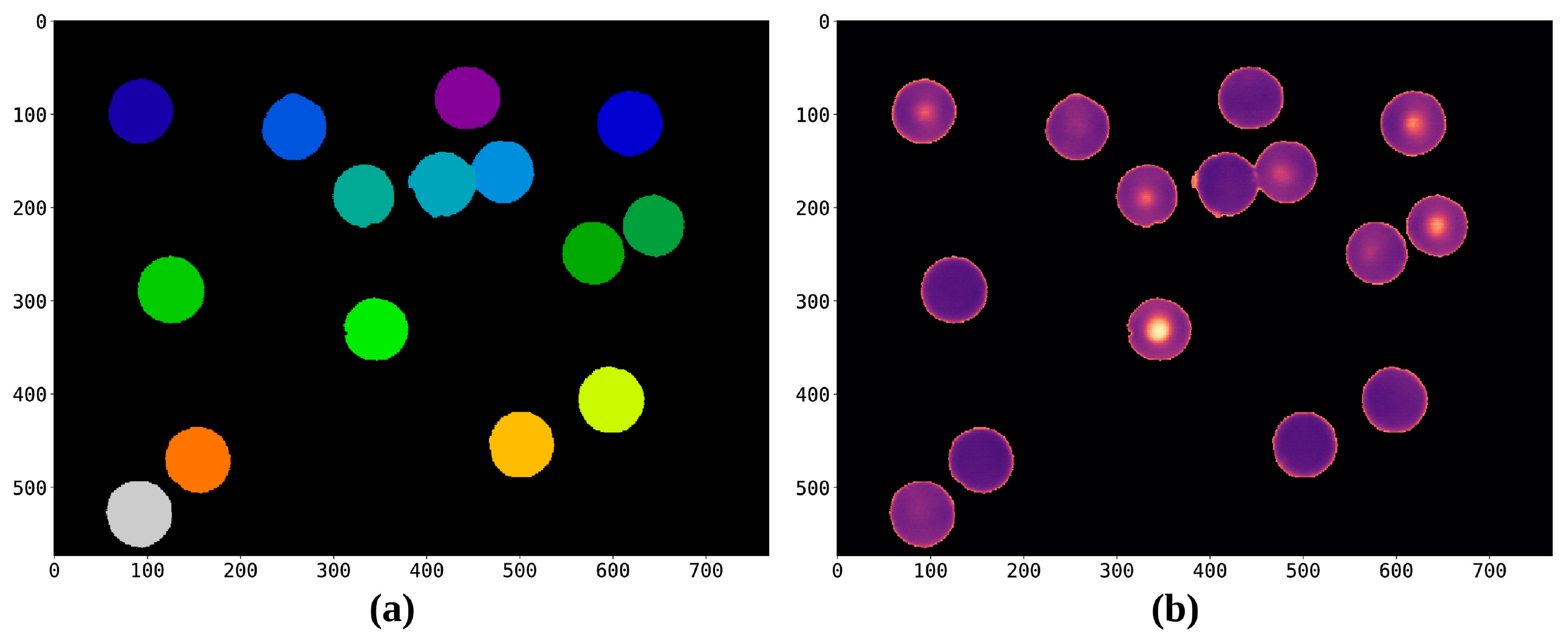}
    \caption{Regions from Figure \ref{fig:test_figure1} complying
    with $\epsilon \leq 0.3$. \textbf{(a)} labeled regions. \textbf{(b)}
    tracks correspondent to (a) in their original gray levels. Colormaps:
    (a) \texttt{nipy\_spectral}. (b) \texttt{magma}.}
    \label{fig:circular_tracks}
\end{figure}

In this application, we use WUSEM to relate the track energies to the
product of major and minor diameters ($D_>$ and $D_<$, respectively) and
the mean gray level of each track. We considered only approximately
circular tracks in our analysis, based on an eccentricity
($\epsilon$)\footnote{In image processing, the eccentricity of an object
is a number in the interval $[0, 1)$. The lower the value, the region
becomes closer to a circle.} criterion: $\epsilon$ should be equal to
or less than 0.3 (Figure \ref{fig:circular_tracks}). This additional
criterion can be used also for counting tracks, to ensure that false
positives (spurious objects counted as tracks) are avoided.

\begin{figure*}[htb]  
    \centering
    \includegraphics[width=1\textwidth]{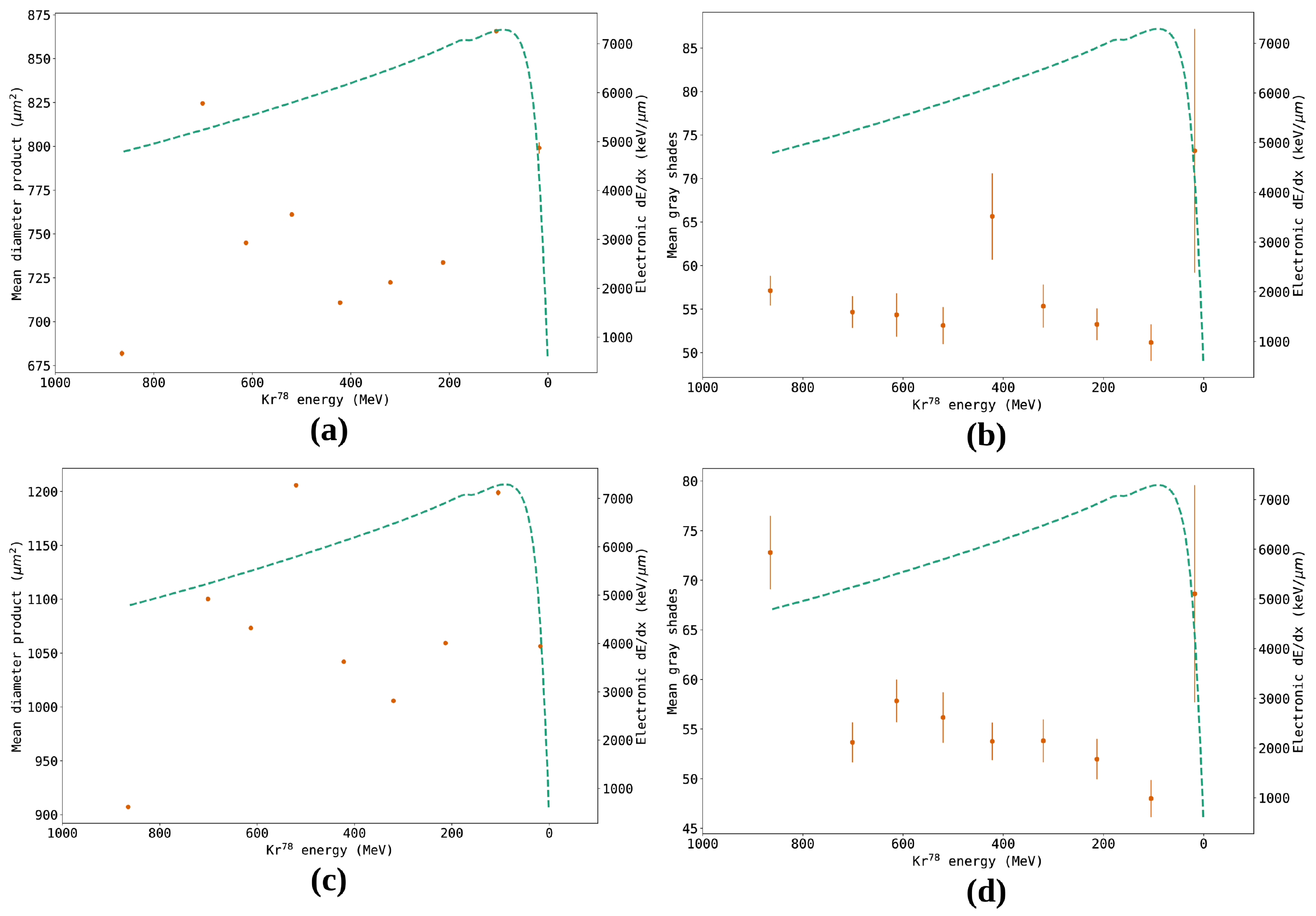}
    \caption{Relation between incident energy versus mean diameter product
    (\textbf{(a)} 4.5 min; \textbf{(c)} 8.5 min samples, left Y axis)
    and incident energy versus mean gray levels (\textbf{(b)} 4.5 min;
    \textbf{(d)} 8.5 min samples, left Y axis). Cyan dashed line:
    electronic energy loss calculated with SRIM (right Y axis).}
    \label{fig:incident_energy}
\end{figure*}

After separating each track, we can obtain its features such as gray
levels and diameters. Once the mean gray level and diameters of each
track in a photomicrograph are obtained, we can relate them with the
incident energy for each sample. The mean gray level of each sample is
obtained getting the mean of all gray levels of the tracks in the images
of that sample. We adopted a similar process to obtain the mean diameters.
Then, the results are related to the incident energy (Figure
\ref{fig:incident_energy}).

The diameter products roughly reflect the electronic energy loss (dE/dx)
curve for $^{78}Kr$ in DAP (Figure \ref{fig:incident_energy} (a) and (c)),
calculated with the software SRIM \cite{ZIEGLER2010}. The Bragg peak appears
around 100 MeV. Further scatter of points can be attributed to poor control
of etching conditions. The uncertainty of 3 $^\circ C$ in temperature may
cause a large variation in etching results \cite{GUEDES1998}. Variations
in gray level means are impaired probably because this set was acquired in
reflected illumination mode, which privileges surface details over depth
effects.

\subsection{Processing photomicrographs of fission tracks in DAP}

In photomicrographs from the first dataset, our main concern was track
superposition. However, all tracks were similar, created by a collimated
beam of $^{78}Kr$ tracks. Here we go one step ahead, applying WUSEM to
images where tracks present a variety of shapes. This image dataset was
obtained from DAP plaques irradiated with thermal neutrons, coupled with
apatite mounts. Fission fragments are born in the interior of the mineral,
then emitted at different directions towards the detector. For instance,
round tracks were created by perpendicular incident fragments, while the
elliptical ones were created by particles hitting DAP surface at shallower
angles (Figure \ref{fig:test_figure2}).

As in the previous analysis, the manual counting performed by an experienced
observer is taken as reference because we expect the observer to be able
to recognize tracks efficiently. However, in this case, we do not expect
the observer efficiency to 100 \%. Fragments hitting the detector at lower
energies originate tracks that are very difficult to distinguish from
detector surfaces. An experienced observer would avoid counting those
tracks to keep hers/his counting efficiency constant.

\begin{figure*}[htb]  
    \centering
    \includegraphics[width=1\textwidth]{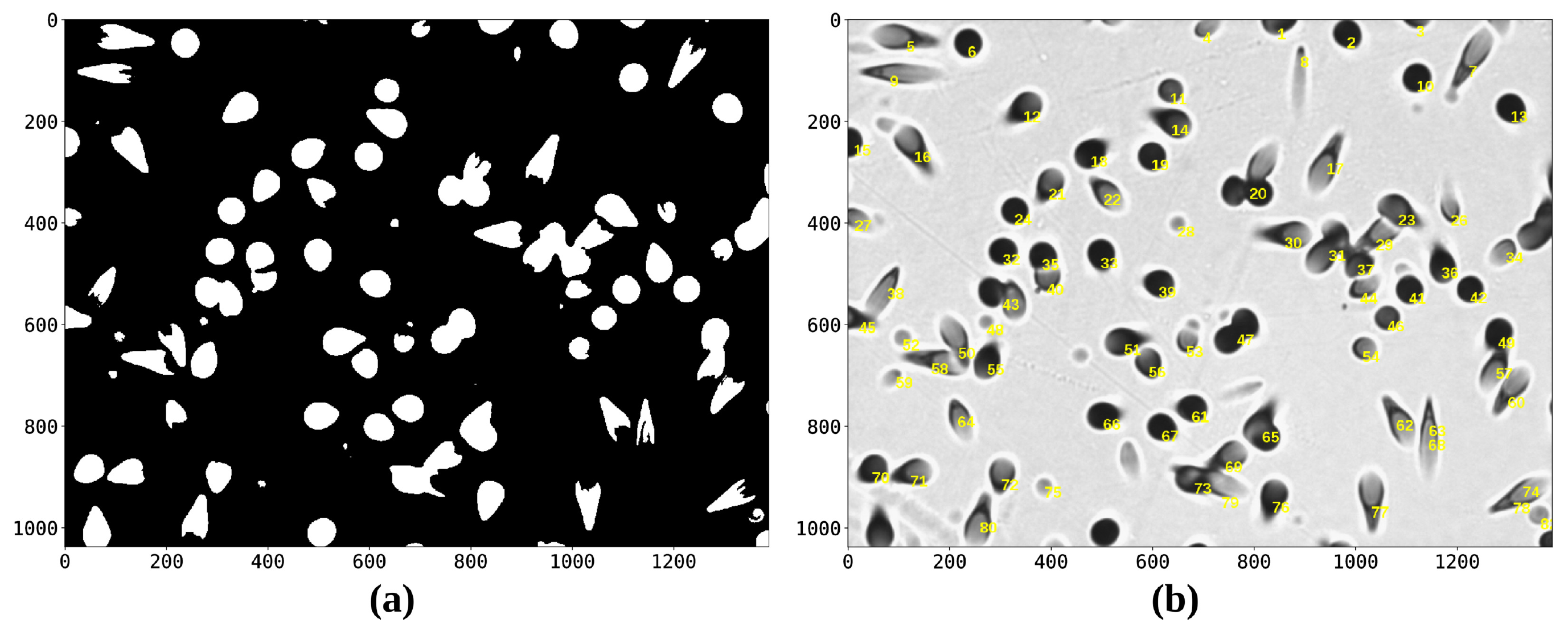}
    \caption{\textbf{(a)} Input photomicrograph (Figure
    \ref{fig:test_figure2}) binarized using the ISODATA threshold
    (threshold = 0.59) and region filling. \textbf{(b)} Tracks separated
    in Figure \ref{fig:test_figure2} using the WUSEM algorithm, and then
    enumerated using the function \texttt{enumerate\_objects()}.
    Parameters for WUSEM algorithm: \texttt{initial\_radius = 5},
    \texttt{delta\_radius = 12}.}
    \label{fig:enumerated_result2}
\end{figure*}

Repeating the previous processes for photomicrographs in the second dataset, we
first binarize a test photomicrograph\footnote{Image ``\texttt{FT-Lab\_19.07.390.MAG1.jpg}'',
from the folder \texttt{orig\_figures/dataset\_02}. Available in the
Supplementary Material.} (Figure \ref{fig:test_figure2}) using
the ISODATA threshold. The binarized image is generated for two scenarios:
considering and ignoring border tracks. Here, regions in the binary image
are also filled using the function
\texttt{ndimage.morphology.binary\_fill\_holes()} from scipy. Then we
apply the WUSEM algorithm. We chose \texttt{initial\_radius = 5} and
\texttt{delta\_radius = 2} as parameters, and the WUSEM result as
parameter in the function \texttt{enumerate\_objects()} (Figure
\ref{fig:enumerated_result2}).

\begin{figure*}[htb]  
    \centering
    \includegraphics[width=0.8\textwidth]{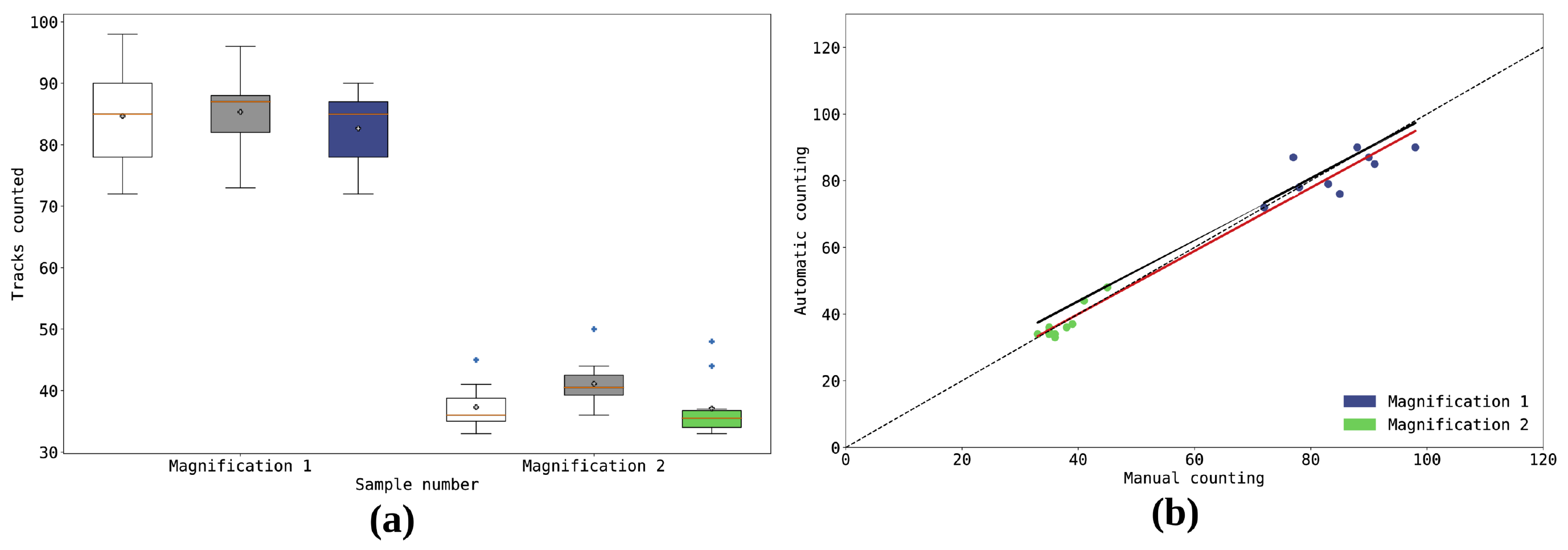}
    \caption{Comparison between manual and automatic counting for
    photomicrographs in dataset 2. \textbf{(a)} white: manual counting.
    Gray: flooding watershed counting. Red line: distribution median.
    White signal: distribution mean. Blue dots: outliers. \textbf{(b)}
    dashed: 1:1 line. Red line: regression for the WUSEM counting data.
    Black line: regression for the flooding watershed counting data.}
    \label{fig:stats_plots2}
\end{figure*}

For this dataset, we established an arbitrary tolerance of five tracks less
than the mean of the manual counting. In this case, WUSEM’s parameters
become candidates if $0 < \mu_{n_{manual}} – n_{auto} < 5$, where $n$ is
the track number obtained by each approach. According to the stated comparison,
the best parameters are \texttt{initial\_radius = 5}, \texttt{delta\_radius = 12}
for the first magnification and \texttt{initial\_radius = 10},
\texttt{delta\_radius = 14} for the second one. Using the best parameters
defined, we compared manual, classic watershed and WUSEM counting for each
sample (Figure \ref{fig:stats_plots2}, Table \ref{tab:stats_set2}).

\begin{table*}[h]  
\centering
    {\scriptsize
    \begin{tabulary}{1.35\textwidth}{CCCC}
    \hline
    \textbf{Magnification} & \textbf{Manual counting ($\mu\pm 1\sigma$)} & \textbf{WUSEM counting ($\mu\pm 1\sigma$)} & \textbf{Efficiency $\pm\,1\sigma$} \\
    \hline
    \textbf{1} & $85 \pm 8$ & $83 \pm 6$ & $0.98 \pm 0.07$ \\
    \textbf{2} & $37 \pm 3$ & $37 \pm 5$ & $0.99 \pm 0.05$ \\
    \hline
    \end{tabulary}
    }
    \caption{Mean, standard deviation, and automatic/manual efficiency
    ratio for each magnification of the second dataset.}
    \label{tab:stats_set2}
\end{table*}

WUSEM succeeded in avoiding false positives in this application, when
compared to the classic watershed (black line above the 1:1 line in Figure
\ref{fig:stats_plots2} (b)). Still, the user could apply more restrictive
criteria. Also, it is worth noting the efficiency variation between the
two image sets. Bigger objects are easier to be treated, thus automatic
track counting in greater magnification images resulted in a higher
counting efficiency (Table \ref{tab:stats_set2}).

\section{Discussion}

\subsection{Structuring elements}

In this study, we used disks as structuring elements for processing images
in both datasets. Since tracks in photomicrographs from dataset 1 are almost
isotropic (as seen in Figure \ref{fig:test_figure1}), disks are
suitable structuring elements to be used in their segmentation. However,
tracks within images in dataset 2 do not have a defined format (Figure
\ref{fig:test_figure2}). Employing different structuring elements,
e.g. rotated cones or ellipses, could improve their segmentation and the
automatic counting result.

\subsection{False positives, false negatives and counting efficiency}

In track counting, reproducibility is not about counting every track in
the image, but counting the same types of tracks every time. It is the
primary concern in FTD: for instance, the efficiencies for counting tracks
in the standard sample should not vary when using zeta age calibrations
\cite{HURFORD1983, HURFORD1990}. For absolute methods of determining neutron
fluence \cite{SOARES2013}, the efficiency should be constant when counting
tracks in unknown age samples. The fast-growing areas of FTD using the Laser
Ablation Inductively Couple Plasma Mass Spectrometer (LA-ICP-MS,
\cite{HADLER2009, HASEBE2004, SOARES2014, SOARES2015}) or the Electron
Microprobe (\cite{GOMBOSI2014, DIAS2017}) have the same efficiency issues,
and could also benefit from automatic counting.

The major challenge for reproducibility is avoiding false positives, spurious
objects such as scratches on the detector or mineral surface and other
etching figures which automatic counting algorithms could misrepresent as
tracks. In most situations, it is preferable to restrict the criteria, thus
increasing the number of false negatives (not counted tracks), even implying
in efficiency reduction.

Even more experienced observers expect some decreasing in efficiency due
to superposition when counting tracks in high track density samples. This
loss tends to be more severe in automatic counting. When applying algorithms
as WUSEM, the separation of tracks in objects formed of several tracks
is not always possible.

Counting less tracks than the actual number in a cluster is acceptable;
however, when processing a large number of clusters in higher track density
samples, we expect a lower efficiency when compared with lower track density
samples. This effect can be assessed by calibrating the efficiency as a
function of track density. Therefore, efficiencies presented for WUSEM
(Tables \ref{tab:stats_set1} and \ref{tab:stats_set2}) only hold for the
track densities of the used samples.

\subsection{Perspective of future development}

\begin{figure*}[htb]  
    \centering
    \includegraphics[width=0.8\textwidth]{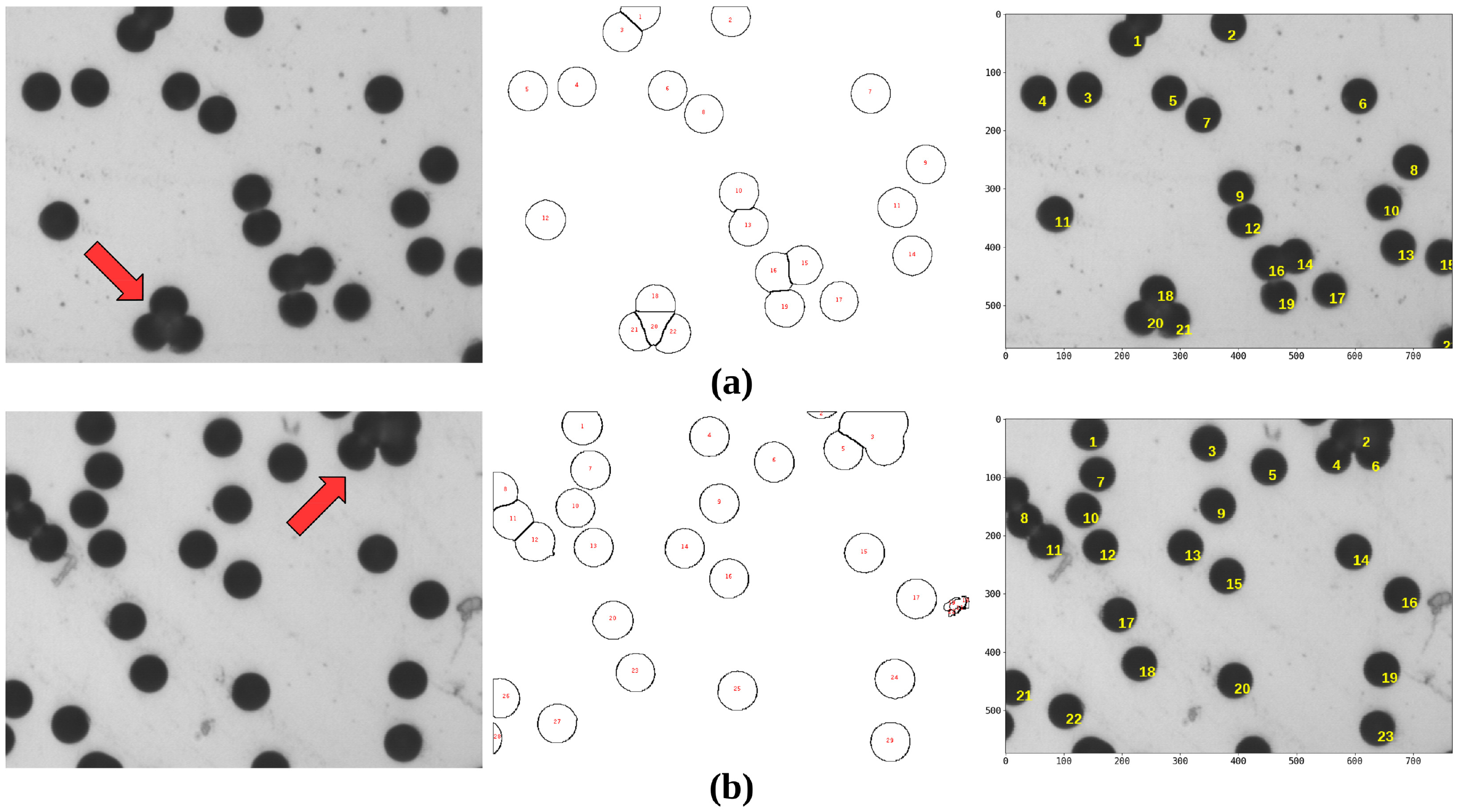}
    \caption{When using suitable input parameters, WUSEM may perform
    better in certain regions where the classic watershed does not
    return reliable results. For instance, the highlighted region in
    \textbf{(a)} presents three tracks. WUSEM separates them correctly,
    but the region is oversegmented by the classic watershed. The
    highlighted region in \textbf{(b)}, by its turn, is undersegmented
    by the classic watershed, which returns two tracks. WUSEM returns
    three tracks, being closer to the real number (four tracks). Left:
    input photomicrographs with highlighted regions. Center: tracks
    separated using WUSEM. Right: tracks separated using classic
    watershed. Parameters for WUSEM algorithm: \texttt{initial\_radius = 15},
    \texttt{delta\_radius = 4}.}
    \label{fig:comparison}
\end{figure*}

The WUSEM algorithm represents an advance in automated track counting,
opening several possibilities. Differently from the direct application
of classic watershed, WUSEM allows adaptation (Figure \ref{fig:comparison}):
the user can determine optimal structuring elements and efficiencies
using a training image subset and apply these parameters to hers/his
sample photomicrographs. This is very convenient in fission track dating,
where the tracks present the same shapes regardless of the track source,
and efficiency may vary mainly due to superposition of tracks in higher
track density samples.

Another possibility is applying WUSEM in separate parts of the input
image, allowing the determination of specific parameters for each track
configuration. This could bring even better results when dealing with
the superposition of two, three or more tracks.

\section{Conclusion}

In this paper we present a watershed algorithm using successive erosions
as markers, which we call WUSEM. We employ WUSEM to separate overlapping
tracks in photomicrographs. WUSEM
performs well in images containing overlapping circular tracks (from a
$^{78}Kr$ collimated beam) and in photomicrographs with fission tracks at
various orientations, both in DAP. The results are encouraging: the mean
automatic/manual counting efficiency ratio is $0.97 \pm 0.11$ when using
WUSEM in the test datasets. We show also that diameter and eccentricity
criteria may be used to increase the reliability of this method.

Since WUSEM using circles as structuring elements is aimed to isotropically
shaped regions, this technique is suitable for separating etched tracks
in DAP. Etching velocity in mineral surfaces depends on the crystal
orientation, yielding more complex etching figures. Also, natural minerals
are richer in scratches and other etching figures that can be mistaken
with fission tracks, especially when using image processing techniques.
WUSEM could be studied to separate tracks in mineral surfaces; for that,
it would need to use different structuring elements, which have to
consider the orientation and shape of each track.

\section*{Acknowledgements}

The authors would like to thank Raymond Jonckheere for the second
photomicrograph dataset and Matthias Schröter for his insights, and
also Christina Trautmann for the sample irradiation at GSI in Darmstadt.
This work is supported by the São Paulo Research Foundation (FAPESP),
grants \# 2014/22922-0 and 2015/24582-4.

\section*{Supplementary Material}

The supplementary material contains the complementary results and the
code published with this study, and is available at
\url{https://github.com/alexandrejaguar/publications/tree/master/2017/dap\_segmentation}.

\section*{References}

\bibliography{afdesiqueira2017_refs}

\end{document}